\documentclass[letterpaper, 10 pt, conference]{ieeeconf}  
\IEEEoverridecommandlockouts
\usepackage{times}
\usepackage{amsmath,amssymb,enumerate}

\usepackage{amsthm}
\usepackage{setspace}
\usepackage{booktabs}
\usepackage{array}
\newcolumntype{P}[1]{>{\centering\arraybackslash}p{#1}}
\usepackage{multicol}
\usepackage[usenames,dvipsnames,svgnames,table]{xcolor}
\usepackage[colorlinks=true,pdfpagemode=UseNone,citecolor=black,linkcolor=black,urlcolor=BrickRed]{hyperref}
\usepackage{mathtools}
\usepackage{graphicx}
\usepackage{caption}
\usepackage{algorithm, algorithmicx, algpseudocode}

\usepackage{blindtext}
\usepackage{gensymb}
\usepackage{xparse}
\usepackage{lipsum}
\usepackage{soul} 
\usepackage{mathrsfs}
\usepackage[mathscr]{euscript}
\usepackage{cite} 
\usepackage[caption=false,font=footnotesize]{subfig}
\usepackage{amsfonts}
\usepackage{cleveref}
\usepackage[utf8]{inputenc}
\usepackage[T1]{fontenc}
\usepackage{textcomp}
\usepackage{balance}
\usepackage{multirow}
\usepackage{makecell} 
\setcellgapes{5pt}

\definecolor{DupontGray}{RGB}{184,171,158}
\definecolor{UMblue}{RGB}{14,43,88}
\definecolor{UMyellow}{RGB}{252,204,6}
\definecolor{VermillionRed}{RGB}{156,31,46}
\definecolor{PullmanGreen}{RGB}{59,51,28}
\definecolor{GNGray}{RGB}{54,54,48}

\newcommand{\m}{\mathop{\mathrm{m}}}

\newcommand{\GHz}{\mathop{\mathrm{GHz}}}
\newcommand{\MHz}{\mathop{\mathrm{MHz}}}

\newcommand{\squeezeup}{\vspace{-3mm}}

\providecommand{\keywords}[1]{\textbf{\textit{Index terms---}} #1}

\title{\LARGE \bf
DeepLocNet: Deep Observation Classification and Ranging Bias Regression for Radio Positioning Systems
}

\author{Sahib Singh Dhanjal, Maani Ghaffari, and Ryan M. Eustice
\thanks{The authors are with the Robotics Institute, University of Michigan, Ann Arbor, MI 48109 USA {\tt\small \{{sdhanjal, maanigj, eustice\}}@umich.edu}.}
}

\begin{document}

\maketitle
\thispagestyle{empty}
\pagestyle{empty}

\begin{abstract}
WiFi technology has been used pervasively in fine-grained indoor localization, gesture recognition, and adaptive communication. Achieving better performance in these tasks generally boils down to differentiating Line-Of-Sight (LOS) from Non-Line-Of-Sight (NLOS) signal propagation reliably which generally requires expensive/specialized hardware due to the complex nature of indoor environments.  Hence, the development of low-cost accurate positioning systems that exploit available infrastructure is not entirely solved. In this paper, we develop a framework for indoor localization and tracking of ubiquitous mobile devices such as smartphones using on-board sensors. We present a novel deep LOS/NLOS classifier which uses the Received Signal Strength Indicator (RSSI), and can classify the input signal with an accuracy of 85\%. The proposed algorithm can globally localize and track a smartphone (or robot) with \emph{a priori} unknown location, and with a semi-accurate prior map (error within $0.8 \m$) of the WiFi Access Points (AP). Through simultaneously solving for the trajectory and the map of access points, we recover a trajectory of the device and corrected locations for the access points. Experimental evaluations of the framework show that localization accuracy is increased by using the trained deep network; furthermore, the system becomes robust to any error in the map of APs. 


\end{abstract}

\section{Introduction}
Low-cost indoor localization solutions using radio signals such as WiFi and Bluetooth have long been studied. Radio signals are easily distorted by the presence of dynamic objects, the room temperature, dust, and even humidity. Furthermore, shadow fading and multipath propagation severely hinder the reliability of signal strength for ranging~\cite{amiot2013pylayers}. The current state-of-art of radio-based positioning techniques~\cite{pap0,pap1} are broadly based on the following four distinct categories: (i)~Received Signal Strength Indicator (RSSI), (ii)~Angle Of Arrival (AOA), (iii)~Time Of Arrival (TOA), and (iv)~Physical Layer Information (PHY). For further details of the above approaches please see~\cite{pap0}. 

\begin{figure}[t]
  \centering
  \includegraphics[width=0.99\columnwidth]{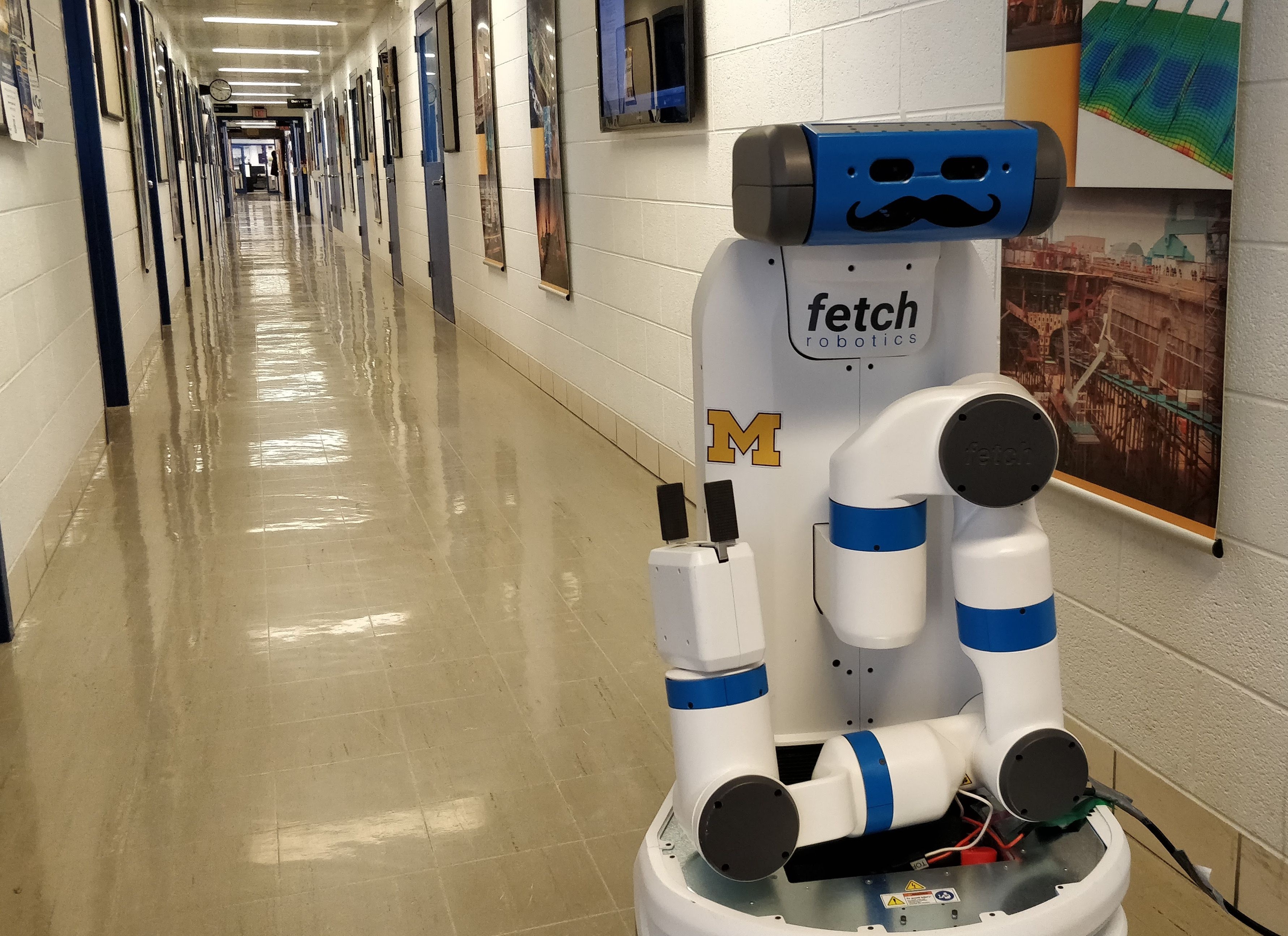}
  \caption{The Fetch Mobile Manipulator and the environment in which we performed the experiments. We manually operated the robot while recording its pose computed using a 2D laser range-finder and the WiFi signatures of all available APs throughout the trial. The recorded poses were used as ground truth and WiFi signatures used as measurements. Since we knew positions of all APs, we manually labeled all data as LOS/NLOS and calculated the Euclidean distances using the floor plan.}
  \label{fig:fetch}
  \squeezeup\squeezeup
\end{figure}

Except (i), all of the above methods require specialized hardware for obtaining range measurements from WiFi Access Points (APs). This requirement limits the applicability of these methods in non-commercial applications. Once range measurements are obtained, positioning techniques such as spherical or hyperbolic positioning can be used to localize the device \cite{pap0}. In order to improve the positioning accuracy, one can combine measurements from multiple sensors such as GPS, magnetometer, and camera using filtering methods such as Kalman filtering or particle filtering \cite{thrun2005probabilistic}. In this work, we are concerned with scenarios where only the RSSI information using a commodity WiFi receiver is available. Such information is ubiquitous and available on even the cheapest smartphones nowadays. We develop a "light-weight" deep neural network for accurately classifying between Line-Of-Sight (LOS) and Non-LOS (NLOS) with the capability of easily being deployed on a smartphone.

The ranging methods mentioned earlier also heavily rely on one of the following two prerequisites about the environment: (i) construction of a radio fingerprint of the entire environment, or (ii) accurate map of the position of each AP. Both of these methods have their associated pros and cons. Radio fingerprinting requires an initial dry run in the environment to record information. The entire area is divided into a grid, and two data points are recorded for each location: (i) the RSSI from each AP, (ii) the standard deviation of each signal. This process is more robust than using a map; however, can be very cumbersome and is not conveniently scalable. Similarly, getting accurate maps of the environment is not always possible.

In this work, a semi-accurate initial map (location of APs within $0.8 \m$ from the ground-truth location) suffices for localization, relaxing the assumption of requiring an accurate initial map. The algorithm estimates the location of the APs, as well as the trajectory of the robot/device. In addition, we introduce a novel deep learning-based LOS/NLOS observation classification and ranging bias regression that is integrated within the widely used particle filtering-based localization and Simultaneous Localization And Mapping (SLAM) frameworks. 


\subsection{Contributions}
The main contributions of this paper are as follows. First, we design a deep network to classify LOS/NLOS signal propagation with $85\%$ accuracy using the passive RSSI information only. The use of passive RSSI information assures that we will not require any custom hardware and the framework will work on any commodity smartphone/robot. The network is designed in a way that it can easily run on embedded devices such as most modern smartphones, without a lot of memory overhead. Second, we use the WiFi sensor model in a position estimation and mapping framework to provide a map of the APs as well as a trajectory of smartphone or robot. By utilizing our approach, the tedious process of WiFi fingerprinting, or the cost of additional specialized hardware is eliminated. Finally, we make the code for the framework, the deep network architecture, and wave propagation simulation environments open source:\\ {\small{\url{https://github.com/sahibdhanjal/DeepLocNet}}}

\subsection{Outline}
A review of related works is given in the following section. Section III describes the problem statement and formulation. An overview of the proposed framework is given in Section IV, followed by experimental methodology and results in Section V. Finally, Section VI concludes the paper by discussing limitations and achievements of the proposed framework and providing ideas for future work.


\section{Related Work}
One of the earliest Radio Frequency (RF)-based localization methods, \emph{RADAR}~\cite{pap9}, used fingerprinting and environment profiling with commodity hardware to provide for indoor localization. Building on top of that, \emph{Horus}\cite{pap8} introduced a client-based probabilistic technique aiming to identify and address channel loss in a light-weight package. Even though accurate up to $0.6 \m$, the major drawback of this method is that localization occurs in two phases: (i) offline map building and clustering phase, (ii) online localization phase.
\par
Similarly, \emph{WiFi iLocate}\cite{pap19} describes a method based on Gaussian process regression and develops a press-to-go package, where an initial run is made for training the model and then online localization takes place. Going a step further, \cite{pap16} tracks footstep data and integrates it with RSSI based range measurements, IMU, and magnetometer to provide an accuracy of up to $0.9 \m$. In \cite{jadidi2017gaussian}, the authors use a Gaussian processing-based method to classify RSSI data from Bluetooth Low Energy (BLE) beacons into LOS/NLOS which is then used in conjunction with an IMU and a particle filter for localization.
\par
Most recent methods have started using the Physical Layer (PHY) information instead of the MAC Layer RSSI information. In \cite{pap7}, the authors devise a new method called \emph{PinLoc}, which is able to localize a device within a $1 \m \times 1 \m$ box. The main observation they made was that dynamic obstructions in the environment can be statistically reproduced. This fact, in turn, was used to detect LOS signals using Bayesian Inference.
\par
A similar method is described in \cite{pap5}, where the authors extract phase, transmission time, and strength information from the PHY layer and incorporate a classifier to localize. Most of the above methods are based on clustering approaches, which limits the capabilities of the system as there is no guarantee that the dataset will inherently contain clusters. To tackle this, the authors in \cite{pap14} came up with a unique statistic called the \emph{Hopkins Statistic}, which measures the clustering tendency to recognize environments. Based on this clustering tendency, they were able to model the environments better, and in turn the location of the APs, leading to improved ranging. However, the method heavily relies on Multiple-Input-Multiple-Output (MIMO) techniques, with a number of antennas, and hence suffers from considerable hardware modification, limiting the practical applications.
\par 

Other methods such as AmpN~\cite{pap13}, leverage the use of amplitude information of the Channel State Information (CSI) from the PHY Layer. These methods measure standard properties such as the \emph{kurtosis}, \emph{Rician K-factor}, \emph{skewness}, and variation~\cite{pap13} among others for each of the CSI amplitudes, and then train a neural network for dynamic classification and recognition of LOS/NLOS signals. The authors of \cite{pap15} add another layer of filtering to better understand the property of LOS/NLOS propagation. In their work, \emph{Bi-Loc}, they use phase errors in conjunction with the the CSI amplitude data, to propose a deep learning approach for fingerprinting. Using this, they have two modalities: Angle of Arrival and CSI, which is then used for fingerprinting. The authors of \cite{pap12} go as far as visualizing the CSI heatmap as an image, and then running a deep convolutional network on it to differentiate between LOS and NLOS.
\par
Although the use of PHY level information has now provided means for more accurate LOS/NLOS classification, and in turn localization, the use of customized hardware \emph{such as Network Interface Cards (NICs)} limits its application to truly mobile and ubiquitous devices (such as smartphones). Some of the issues, identified by these works, to enable real-time LOS identification are as listed below:

\begin{itemize}
\item Commodity WiFi devices fail to support precise Channel Impulse Response measurements due to limited operating bandwidth.
\item Existing channel statistics-based features require large amount of samples, impeding real-time performance.
\item Most LOS identification schemes are designed for stationary scenarios. Even those incorporating slight mobility fail in truly mobile cases.
\item Requirement of custom hardware limits the application of these algorithms for truly mobile cases.
\end{itemize}

In this work, we bring the advances in SLAM to efficiently solve the indoor localization and tracking problem using sensors available in commonly used mobile devices. The main features that distinguishes this work from the available radio signal-based indoor positioning literature are as follows. We develop a deep neural network that can classify between LOS/NLOS using only RSSI information. Our framework also does not require the tedious process of fingerprinting (site survey), hence is more scalable. Moreover, the system is robust to errors ($\leq  0.8m$) in the map of access points due to the usage of FastSLAM to localize the device, as well as obtain a map of the environment.

\section{Problem Formulation and Preliminaries}
We now define the problems we study in this paper and then briefly explain required preliminaries to solve these problems. Let $x_t \in \mathbb{R}^3$ be the device position at time $t$. The device is initially located at $x_0$ which is unknown and can only receive the RSSI of a broadcasted signal. Let $Z_t$ be the set of possible range measurements obtained from converting RSSI at time $t$. Given the set of known APs, we wish to solve the following problems: 
\par
\textit{Problem 1 (Measurement Model): }The measurement model $p(z_t|x_t)$ is a conditional probability distribution that represents the likelihood of range measurements. We want to find the mapping from signal, $s_t$, to range measurements, $z_t$, and the likelihood function describing the measurement noise. 
\par
\textit{Problem 2 (Positioning): }Let $z_{1:t} = \{ z_1,\dots, z_t\} $ be a sequence of range measurements up to time $t$. Let $x_t$ be a Markov process of initial distribution $p(x_0)$ and transition model $p(x_t|x_{t-1})$. Given $p(z_t|x_t)$, estimate recursively in time the posterior distribution $p(x_{0:t}|z_{1:t})$.
\par
\textit{Problem 3 (Access Point Locations): }Let $M = \{ m^{[j]}|j = 1,\dots, n_m \}$ be a set of unknown and partially observable features whose elements, $m^{[j]} \in \mathbb{R}^3$ represent WiFi access point locations with respect to a global frame of reference. Given $p(z_t|x_t)$ and $p(x_{0:t}|z_{1:t})$ recursively estimate $p(m^{[j]} | x_t, z_t)$.
\par
In the first problem, we try to characterize the received signal, and through an appropriate model, transform it to a range measurement. Furthermore, we need to find a likelihood function that describes the measurement noise. The second problem can be seen as a range-only self-localization problem. Finally, the last problem is to estimate the locations of the access points given the location of the device. We now state the main assumptions we use to solve the defined problems:
\par
\textit{Assumption 1 (Known Data Association): }Each access point has a unique hardware identifier that is available to the receiver device. This assumption is usually satisfied in practice as each device has a unique MAC-address that is broadcasted together with the RSSI.
\par
\textit{Assumption 2 (only RSSI available):} We assume that the only available information to the receiver is the RSSI. This is the common case for existing wireless routers and commercial NICs. 
\par

\subsection{WiFi Technology}
WiFi is a technology for radio wireless local area networking of devices based on the IEEE 802.11 standards. It most commonly operates on the $2.4 \GHz$ Ultra-High Frequency (UHF) and $5.8 \GHz$  Super-High Frequency (SHF)  Industrial, Scientific and Medical (ISM) radio bands. These wavelengths work best for line-of-sight. Many common materials absorb or reflect them, which further restricts the range, but can minimize interference between different networks in crowded environments.

\begin{figure}[t]
  \centering
  \includegraphics[width=0.99\columnwidth]{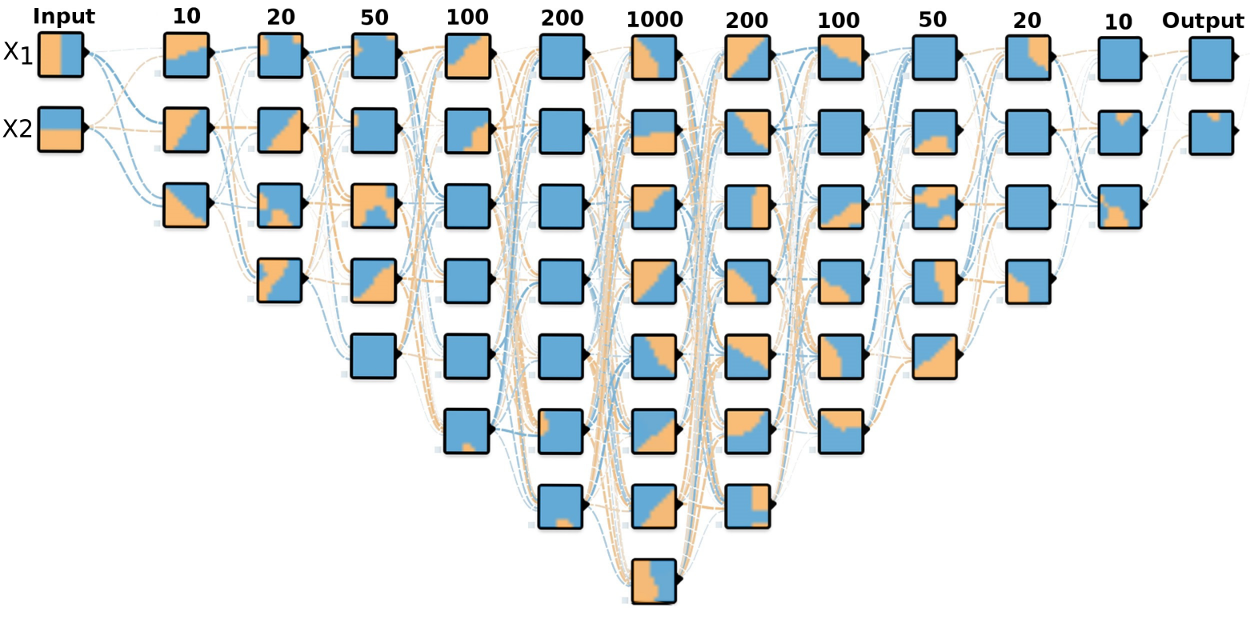}
  \caption{Neural Network Architecture. $x_1$ and $x_2$ represent the Euclidean distance and the distance calculated using Free Space Path Loss (FSPL)~\cite{pap1} formula, respectively. ReLU \cite{agarap2018deep} activation is used between each of the hidden layers. As the network depth increases, it learns the underlying representation of the data, in turn compressing it to learn the underlying structure. The final layer consists of two neurons and the SoftMax activation which gives the probabilities of LOS/NLOS signal propagation as output.}
  \label{fig:nn}
  \squeezeup
\end{figure}

\subsection{Particle Filtering}
In the problem of localization using RSSI, the observation space is nonlinear, and the posterior density is often multimodal. Particle filters are a non-parametric implementation of the Bayes filter that are suitable for tracking and localization problems where dealing with global uncertainty is crucial. Hence, in this work, we use a particle filter and its extension FastSLAM to solve the positioning and mapping problems~\cite{thrun2005probabilistic,jadidi2017gaussian,GhaffariJadidi2018}.


\section{The DeepLocNet Framework}
The proposed framework can be divided into two distinct parts as follows:

\subsection{LOS/NLOS Classifier} The first part of the framework comprises of a deep neural network which is trained to classify between LOS and NLOS signal propagation. The network includes a total of 12 fully connected (FCN) layers which first expand and then contract. The architecture of the network we are using can be seen in Fig.~\ref{fig:nn}. The structure of the neural network is inspired by an autoencoder, where we learn the low-level features representing the underlying data by scaling up, and then scaling down to compress these low-level features into output dimensions. The only difference here is instead of using the input as the output, we are using the class as the output to compute classification loss. \texttt{ReLU}~\cite{agarap2018deep} activation is used between every layer and is found out to work better than \texttt{tanh} activation during the training and experimentation phases. One of the reasons is that the \texttt{tanh} activation function causes saturation of multiple neurons during the training process. The network is trained using the cross entropy loss function given by 

\begin{equation}
H(y,\hat{y}) = -y \log(\hat{y}) - (1-y) \log(1-\hat{y}),
\end{equation} 
where $y$ is the probability of the true label being $1$ \emph{(that is, $p_{\text{label}=1} = y$)} and $\hat{y}$ is the probability of the label predicted by the network being $1$. Cross entropy was chosen over other losses such as MSE or Hinge loss as they are mathematically ill-defined for the classification problem. Before finalizing the network architecture, we tried several different architectures to learn a valid representation of the signal propagation. We also tried going deeper than the current depth, but that led to overfitting on the training data, causing a substantially less testing accuracy than what we obtained using this network. Another reason going deeper did not help is that the number of parameters in the network increased without a substantial increase in classification accuracy.
\par
The number of neurons used per hidden layer was decided based on trial-and-error and the complexity of the signal propagation we wanted to represent. The input data is highly non-linear and signal propagation with shadowing and multi-path effects cause further non-linearity. Because of this complexity, we increased the number of hidden layers until we obtained an increase in both training and testing accuracy. A summary of a few of the well-performing architectures, number of network parameters, and classification accuracies are shown in Table~\ref{table:archs}.

\begin{table}[t!]
    \centering
    \resizebox{\columnwidth}{!}{
    \begin{tabular}{cccc}
        \rowcolor{UMblue!30} \multicolumn{4}{c}{\textbf{Neural Network Configurations}} \\
        \rowcolor{UMblue!30} \textbf{A} & \textbf{B} & \textbf{C} & \textbf{D} \\
        \rowcolor{UMblue!30} \textbf{5 weight layers} & \textbf{8 weight layers} & \textbf{9 weight layers} & \textbf{12 weight layers} \\
        \midrule
        \rowcolor{UMyellow!30} \multicolumn{4}{c}{Input Layer (2 x 1 matrix)} \\
        \midrule
        FCN(2,10) & FCN(2,10) & FCN(2,10) & FCN(2,10) \\
        \midrule
        \rowcolor{UMyellow!30} \multicolumn{4}{c}{ReLU Layer} \\
        \midrule
        FCN(10,20) & FCN(10,20) & FCN(10,20) & FCN(10,20) \\
        \midrule
        \rowcolor{UMyellow!30} \multicolumn{4}{c}{ReLU Layer} \\
        \midrule
        FCN(20,50) & FCN(20,50) & FCN(20,50) & FCN(20,50) \\
        \midrule
        \rowcolor{UMyellow!30} \multicolumn{4}{c}{ReLU Layer} \\
        \midrule
        FCN(50,100) & FCN(50,100) & FCN(50,100) & FCN(50,100) \\
        \midrule
        \rowcolor{UMyellow!30} \multicolumn{4}{c}{ReLU Layer} \\
        \midrule
        FCN(100,2) & FCN(100,50) & FCN(100,200) & FCN(100,200) \\
        \cline{2-4}
        \rowcolor{UMyellow!30} & \multicolumn{3}{c}{ReLU Layer} \\
        \cline{2-4}
         & FCN(50,20) & FCN(200,500) & FCN(200,1000) \\
        \cline{2-4}
        \rowcolor{UMyellow!30} & \multicolumn{3}{c}{ReLU Layer} \\
        \cline{2-4}
         & FCN(20,10) & FCN(500,1000) & FCN(1000,200) \\
        \cline{2-4}
        \rowcolor{UMyellow!30} & \multicolumn{3}{c}{ReLU Layer} \\
        \cline{2-4}
         & FCN(10,2) & FCN(1000,2000) & FCN(200,100) \\
        \cline{3-4}
        \rowcolor{UMyellow!30} & & \multicolumn{2}{c}{ReLU Layer} \\
        \cline{3-4}
        & & FCN(2000,2) & FCN(100,50) \\
        \cline{3-4}
        \rowcolor{UMyellow!30} & & & ReLU Layer\\
        \cline{4-4}
        & & & FCN(50,20) \\
        \cline{4-4}
        \rowcolor{UMyellow!30} & & & ReLU Layer \\
        \cline{4-4}
        & & & FCN(20,10) \\
        \cline{4-4}
        \rowcolor{UMyellow!30} & & & ReLU Layer \\
        \cline{4-4}
        & & & FCN(10,2) \\
        \midrule
        \rowcolor{UMyellow!30} \multicolumn{4}{c}{SoftMax Layer} \\
        \midrule
        \rowcolor{UMblue!30} \multicolumn{4}{c}{\textbf{Classification Accuracies}} \\
        $63.24\%$ & $79.65\%$ & $73.88\%$ & $\mathbf{85.25\%}$ \\
        \midrule
        \rowcolor{UMblue!30} \multicolumn{4}{c}{\textbf{Number of Parameters}} \\
        $182$ & $262$ & $3882$ & $1762$ \\
        \bottomrule 
    \end{tabular}%
    }
    \caption{Neural Network Architectures. FCN(x,y) represents a Fully Connected layer taking x inputs and mapping it to y outputs. ReLU Layer is the ReLU activation \cite{agarap2018deep} function and SoftMax Layer represents the SoftMax activation function.}
    \label{table:archs}
    \squeezeup
\end{table}

\subsection{Localization Framework} 
The localization framework is responsible for the positioning of the device given an estimate of the motion and the received signal strengths of the WiFi signal. We divide this framework into two categories as follows.
\subsubsection{Map accurately known} In the case where the map is accurately known, we use a particle filter \cite{thrun2005probabilistic} to estimate the position of the robot/device. The Sample Importance Resampling (SIR) particle filter we use consists of three main modules: (i) Motion model (sample), (ii) Measurement model (importance), and (iii) resampling. 
\par
\textbf{Motion model} is responsible for generating a set of hypothesis for the current position, based on the previous position and the action taken. More specifically, it specifies a probability $p(x_t | x_{t-1},u_t)$, that action $u_t$ carries the robot from state $x_{t-1}$ to $x_t$ . Let the number of particles generated be $n_{p}$, then the motion model generates a position hypothesis for each of the particles. 
\par
\textbf{Measurement model} is responsible for assigning the weights (or \emph{importance weights}) to each of the particles sampled from the motion model. Since we only obtain range measurements from the sensor, the measurement function can be given as the distance $z_t$ between the current position $x_t$ and the location $m^{[j]}$ of the $j^{th}$ access point as calculated using the Free Space Path Loss (FSPL)~\cite{pap1} equation. Hence, our measurement model using only WiFi can be given as follows: 
\begin{equation}
    d_{Euc}(x_t, m^{[j]}) = \left( (x_{t} - m^{[j]})^{\mathsf{T}} (x_{t} - m^{[j]}) \right)^{\frac{1}{2}}
    \label{deuc}
\end{equation}
\begin{equation}
    d_{rssi}(RSSI^{[j]}) = \frac{1}{20} 10^{\lvert RSSI^{[j]} \rvert - K - 20 \log_{10}(f)},
    \label{drssi}
\end{equation}
where $d_{Euc}$ is the Euclidean distance between the current position $x_t$ and the position $m^{[j]}$ of the $j^{th}$ access point. Similarly, $d_{rssi}$ is the distance of the $j^{th}$ access point based on the RSSI value that the device receives at position $x_t$. $f$ is the frequency of the signal in $\MHz$. $K$ is a constant that depends on the units for $d_{rssi}$ and $f$. For $f$ in $\MHz$ and $d$ in $\mathrm{km}$, $K=32.44$ \cite{amiot2013pylayers}.%
\par

\begin{algorithm}[t]
  \caption{\texttt{DeepLocNet-Measurement-Model}}
  \footnotesize
  \begin{algorithmic}[1]
  \Require Set of particles $\{X_t, w_t\}$ sampled from motion model, $RSSI_j$ from each AP $m^{[j]}$, measurement noise variance $sz_j$ for each AP $m^{[j]}$;
    \For{$i \in X_t$}
        \For{$j \in M$}
            \State $\sigma_n \gets 3$ 
            \State $w_{total} \gets 0$ 
            \State $de \gets d_{Euc}(x_t^{[i]}, m^{[j]})$ \Comment{Euclidean distance}
            \State $dr \gets d_{rssi}(RSSI_j)$ \Comment{RSSI distance}
            \If{$de \leq R$} \Comment{$R$ is the sensing range of the device}
                \If{use classifier}
                    \If{hard classification}  \Comment{hard classification case}
                    \State $label \gets \texttt{getLabel}(de, dr)$
                    \If{$label ==$ \texttt{LOS}}
                    \State $dz \gets \lvert dr-de \rvert$
                    \EndIf
                    \Else  \Comment{soft classification case}
                    \State $p_{LOS}, p_{NLOS} \gets \texttt{getProbs}(de, dr)$ 
                    \State $d_{LOS} \gets \lvert dr-de \rvert$
                    \State $d_{NLOS} \gets \lvert dr - dn \rvert$ \Comment{$dn \sim \mathcal{N}(R,\sigma_n^2)$}
                    \State $dz \gets p_{LOS} \cdot d_{LOS} + p_{NLOS} \cdot d_{NLOS}$
                    \EndIf
                \Else  \Comment{no classification case}
                    \State $dz \gets \lvert dr-de \rvert$
                \EndIf
            \EndIf
        \EndFor
        \State $w_t^{[i]} \gets w_t^{[i]} \cdot f(dz; 0, sz)$ \Comment{Gaussian likelihood $f(x;\mu, \sigma^2)$}
        \State $w_{total} \gets w_{total} + w_t^{[i]}$
    \EndFor
    \For{$i \in X_t$} \Comment{weight normalization}
        \State $w_t^{[i]} \gets w_t^{[i]}/w_{total}$
    \EndFor
    \State \Return $\{X_t, w_t\}$
  \end{algorithmic}
  \label{measure_model}
\end{algorithm}

We implement three methods in the measurement model: (i) No Classification (NC), (ii) Hard (acceptance/rejection) Classification (HC), and (iii) Soft (probabilistic) Classification (SC). The first case represents the naive implementation of the measurement model. In the second case, we only consider LOS signal propagation for ranging (hence hard classification), whereas in the third case, we use probabilities of the signal being LOS or NLOS to calculate an importance weight for the particle. 
For soft classification, $\sigma_n$ can be given as the standard deviation in the maximum range the device is able to sense. We assign it an arbitrary value of $3$ assuming that the changes in power transmission would not change the distance ($d_{Euc}$) more than $3 \mathrm{m}$. We use two functions which access the classifier, \texttt{getLabel()} and \texttt{getProbs()}. Both of them take the Euclidean distance and RSSI based distance using FSPL as inputs. While \texttt{getLabel()} returns the label predicted by the network, \texttt{getProbs()} returns the probabilities that the given inputs are either LOS or NLOS. The overall measurement model we use for the particle filter is as given in Algorithm \ref{measure_model}. 
\par

\textbf{Importance Sampling} draws, with replacement, $n_{p}$ particles from the set $X_t$ of generated and weighted particles using the above two steps. The probability of drawing each particle is given by its importance weight. The resampling essentially transforms the particle set of size $n_{p}$ into another particle set of the same size by replicating particles with higher weights and, in the end, setting all weights uniformly. The resulting sample set usually possesses many duplicates, since particles are drawn with replacement.

\subsubsection{Map partially known} Particle filtering is not effective when the locations of the access points are partially known. Since we are not sure of the access point locations, ranging using the measurement model, as discussed in the previous section, becomes inaccurate. To tackle such cases, the FastSLAM~\cite{thrun2005probabilistic} algorithm is used to provide us with an effective means of localization, as well as a method of rectifying the locations of the access points using the sensor measurements.

The FastSLAM algorithm, in essence, is a particle filter where each particle comprises a map of the locations of each detected access point in addition to the weight and the position of the device. The access point locations are tracked using an Extended Kalman Filter, whereas the robot position is tracked using Particle Filtering. Analogous to the previous case, we divide the measurement model into 3 cases here as well. We refer the reader to \cite{thrun2005probabilistic} for implementation and other details.

\begin{table}[t]
\centering
\resizebox{\columnwidth}{!}{
\begin{tabular}{cccc}
    \toprule 
    \multicolumn{4}{c}{\textbf{2D Localization for the Office Environment}} \\
    \midrule 
    Algorithm &  Classifier (Y/N) & Hard/Soft Class. (H/S) & Loc. RMSE (in m)\\
    \midrule
    PF & N & N/A & 6.3581 \\
    PF & Y & H & 1.8491 \\
    PF & \textbf{Y} & S & \textbf{1.5912} \\
    FS & N & N/A & 6.9249 \\
    FS & Y & H & 1.9071 \\ 
    \textbf{FS} & \textbf{Y} & S & \textbf{1.8169} \\
    \midrule
    \multicolumn{4}{c}{\textbf{3D Localization for the Office Environment}}\\
    \midrule 
    Algorithm &  Classifier (Y/N) & Hard/Soft Class. (H/S) & Loc. RMSE (in m)\\
    \midrule
    PF & N & N/A & 8.6115 \\
    PF & \textbf{Y} & H & \textbf{2.2447} \\
    PF & Y & S & 2.4827 \\ 
    FS & N & N/A & 10.7775 \\
    FS & Y & H & 3.5645 \\
    \textbf{FS} & \textbf{Y} & S & \textbf{3.3193} \\
    \bottomrule 
\end{tabular}
}
\caption{Results for 2D/3D Simulation Experiments in the Office Environment for a single trial. We use 2 algorithms - (i) Particle Filter (PF), (ii) FastSLAM (FS) with the measurement models as described in the DeepLocNet Framework. For each experiment we calculate the RMSE (root mean square error) between each of the waypoints of the ground truth and the localized path.}
\label{tab:results}
\end{table}

\begin{figure}[t]
  \centering
   \subfloat[]{\includegraphics[width=0.5\linewidth]{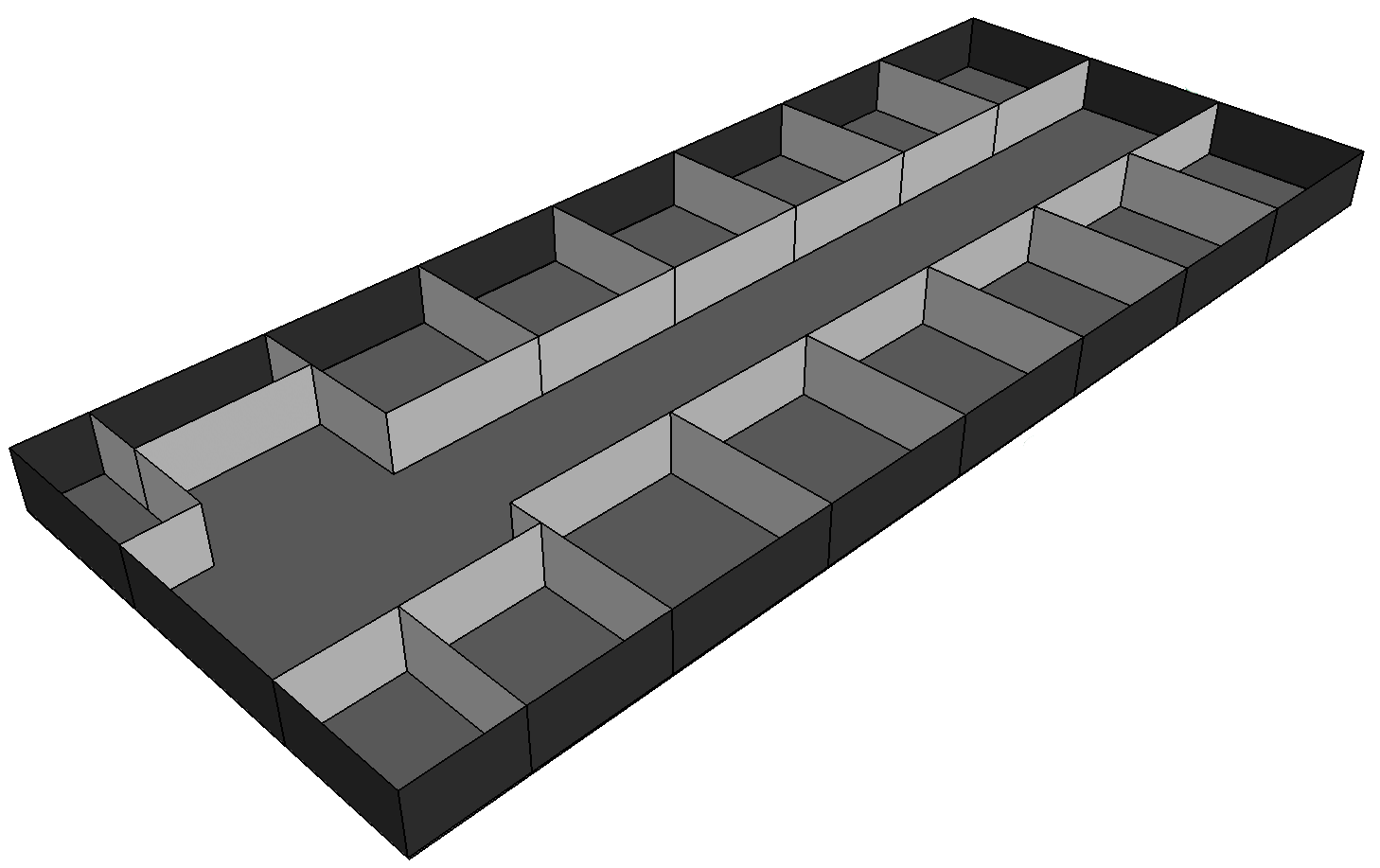}}
   \subfloat[]{\includegraphics[width=0.5\linewidth]{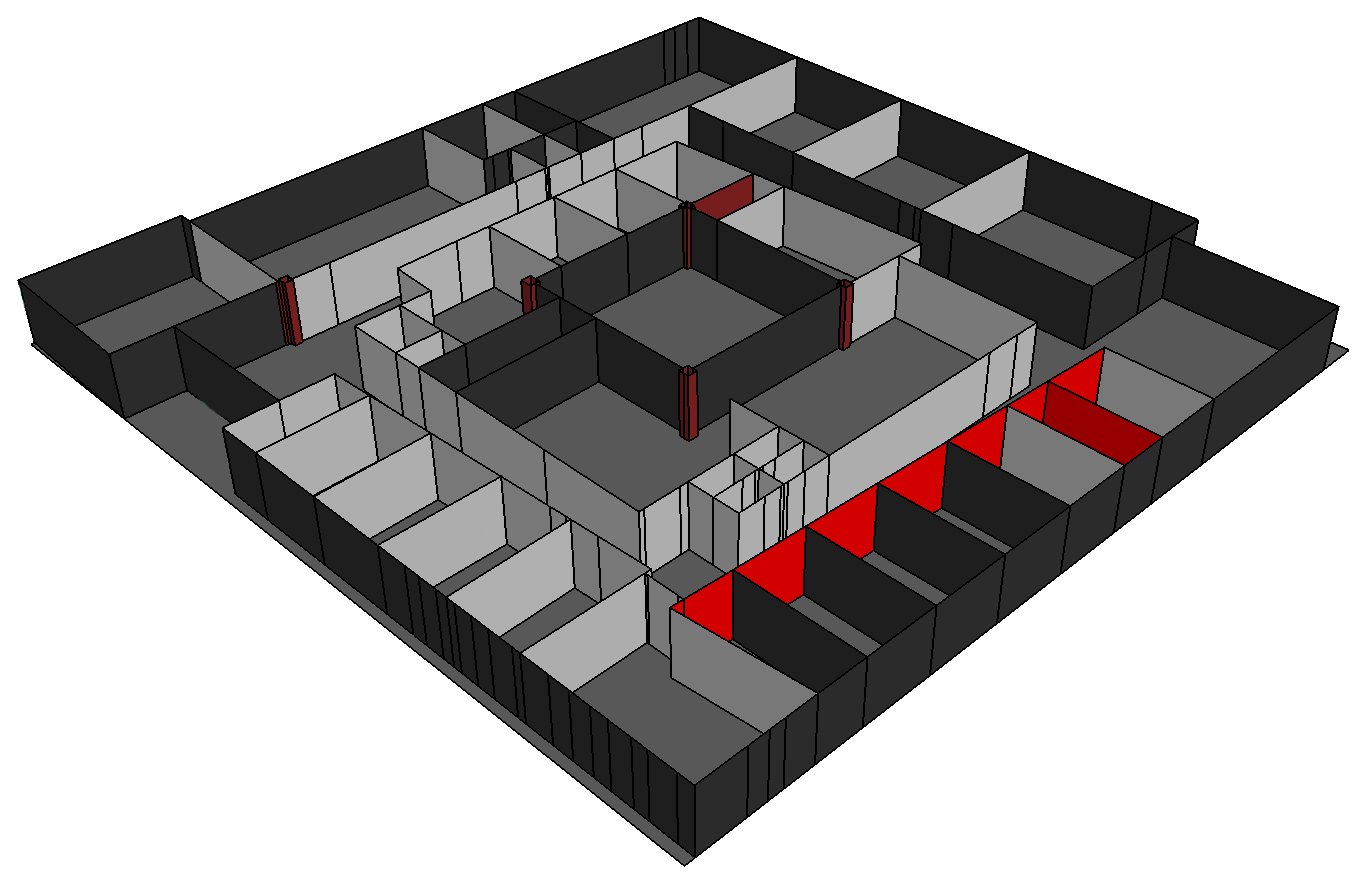}}
  \caption{\textit{(top to bottom)} Environments\cite{amiot2013pylayers}: (a) Office ($52m$ x $9.5m$), (b) W2PTIN ($49m$ x $49m$) }
  \label{fig:envs}
\end{figure}

\section{Experimentation and Evaluation}
In this section, we define our experimentation apparatus and methods. We divide this section into the following subsections:

\subsection{Deep Neural Network Training: } 

\begin{figure}[t]
  \centering
  \subfloat[]{\includegraphics[width=.45\linewidth]{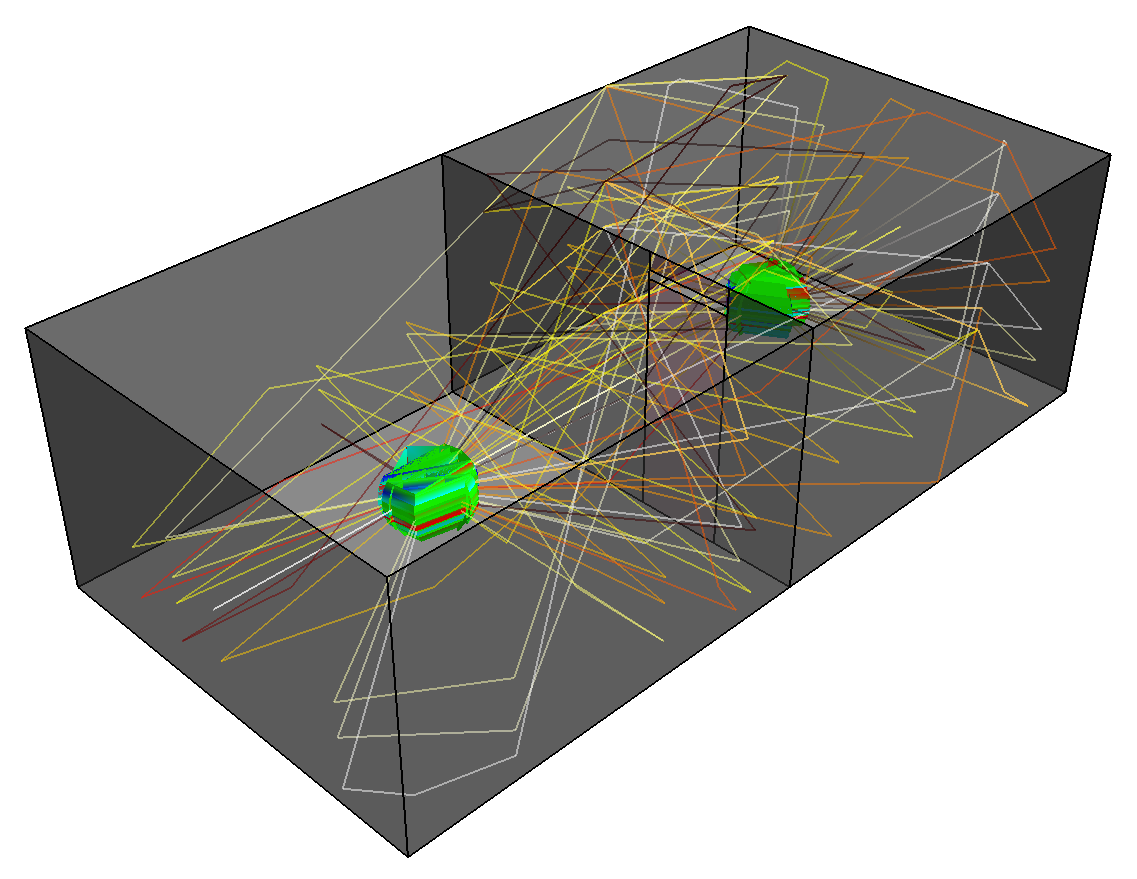}}
  \subfloat[]{\includegraphics[width=.55\linewidth]{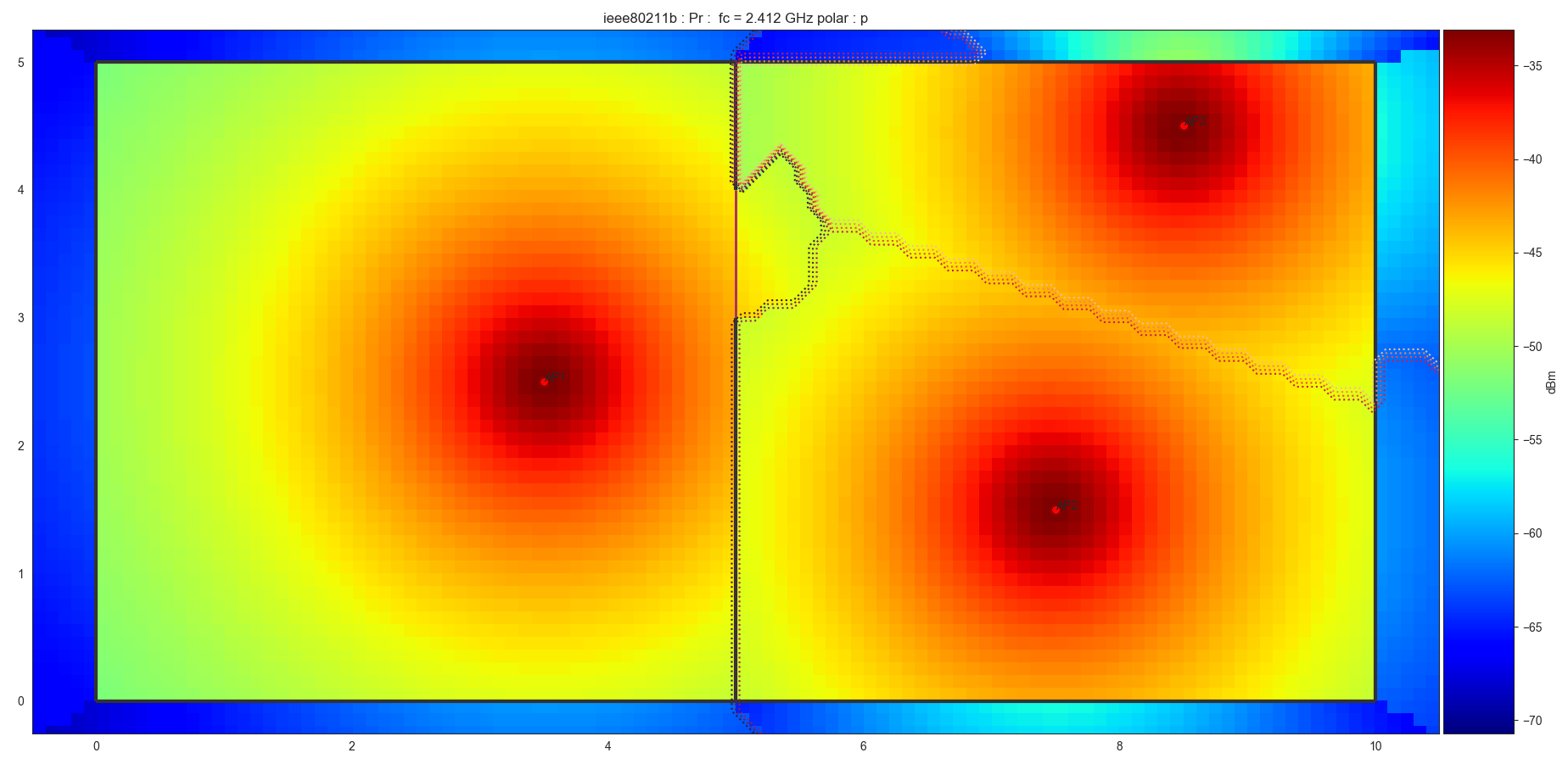}}
  \caption{(a) Simulated wave propagation and (b) Motley Keenan Path Loss in the Defstr ($13m$ x $8m$) environment with 2 and 3 access points respectively.}
  \label{fig:defstr}
\end{figure}

The focus of this part was to collect data to train our network. The training was done in 2 phases: \textit{(i)} initial training on simulated data, and \textit{(ii)} network weight refinements on real-world data. Since obtaining floor plans and blueprints with access point locations may not always be a feasible task, we used \textit{Pylayers} \cite{amiot2013pylayers}, an open-source tool, to simulate wave propagation in complex and dense environments. Using this tool, we defined 30+ environments with varying temperatures, humidity, AP locations, AP characteristics (transmission power, antenna directions, etc) and floor plans. Some of the environments are shown in Fig. \ref{fig:envs}. Results of Motley Keenan Path Loss and 3D Ray tracing for one of the environments can be seen in Fig. \ref{fig:defstr}.
\par

A random walk algorithm was implemented which was able to navigate through a given environment in three dimensions given random start and goal points. We used this algorithm to generate waypoints and calculate the RSSI, Euclidean distance, and label (LOS/NLOS) of each AP at each waypoint in the path. We used Bresenham's line algorithm to calculate if there was any obstruction between the said AP and the current waypoint. If there is an obstruction, we label it as NLOS, otherwise LOS. We simulated over 10 million data points and initially trained our neural network on this data. 
\par
We also had access to the blueprint of one of the buildings on our campus. We implemented software which aided us in data-collection given these blueprints as the robot was moved through the environment. We collected data from this building and used it to refine the weights of our deep neural network.
\par

From the several architectures we tested, network D in Table \ref{table:archs} performed the best both in simulation as well as hardware experiments. The classifier was able to classify LOS from NLOS signal propagation with an accuracy of $\approx 85\%$. From multiple experiments, we found out that the maximum misclassification occurs when the ground truth is NLOS but the prediction is LOS. The reason for this happening is that in cases when the device is near the access point (distance $<10$ map units or $2.5 m$), the signal propagation might be NLOS because of the presence of an obstruction in between the AP and the device (for instance a wall), the Euclidean distance and the distance calculated using the path-loss formula is approximately the same. Hence, though the signal is NLOS, it is predicted as LOS.

\subsection{Experimentation in Simulation: } 
\begin{figure}[t]
  \centering
  \includegraphics[width=0.75\linewidth]{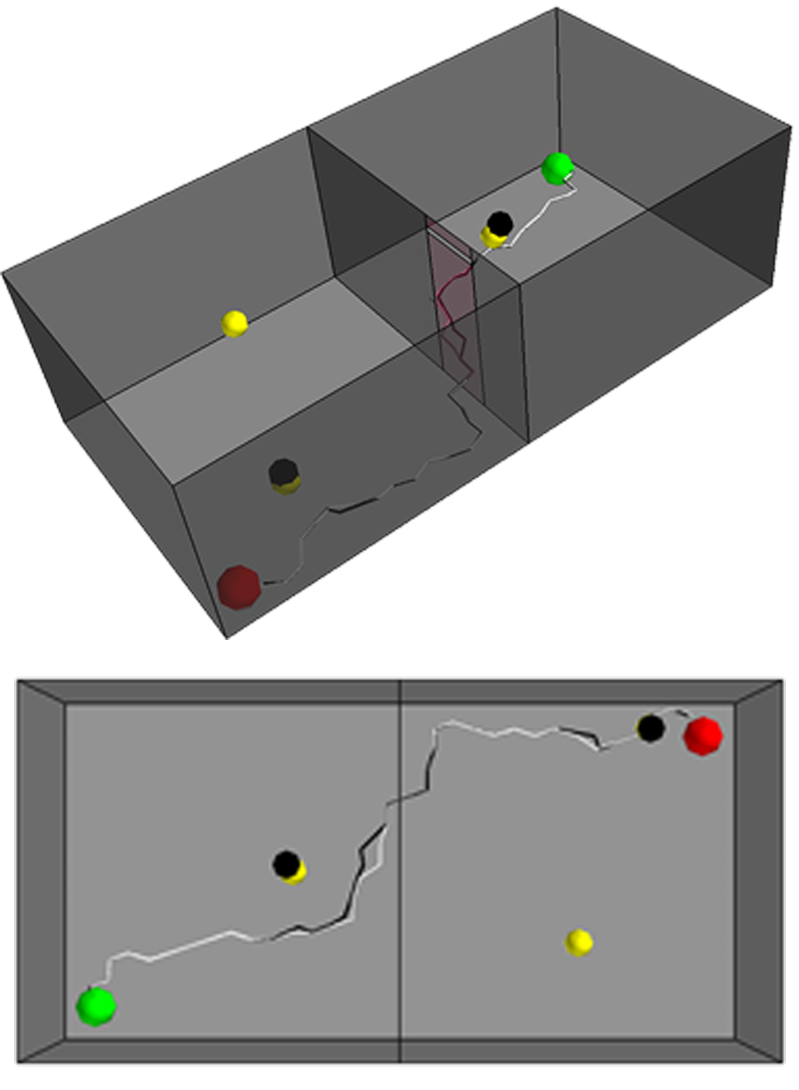}
  \caption{3D Localization in the Defstr environment\cite{amiot2013pylayers}  using FastSLAM and soft classification (top to bottom - orthographic view, top view). The yellow spheres represent actual location of access points, black spheres the localized estimates of the access points, the white pipe the ground truth path and the black pipe the localized path. The green and red spheres represent the start and goal positions respectively. One AP is not detected in this case because it is out of sensing range (2m).The red wall is a door, hence the device is allowed to pass through it.}
  \label{fig:3dlocfs}
\end{figure}

\begin{figure}[t]
  \centering
  \includegraphics[width=0.95\columnwidth]{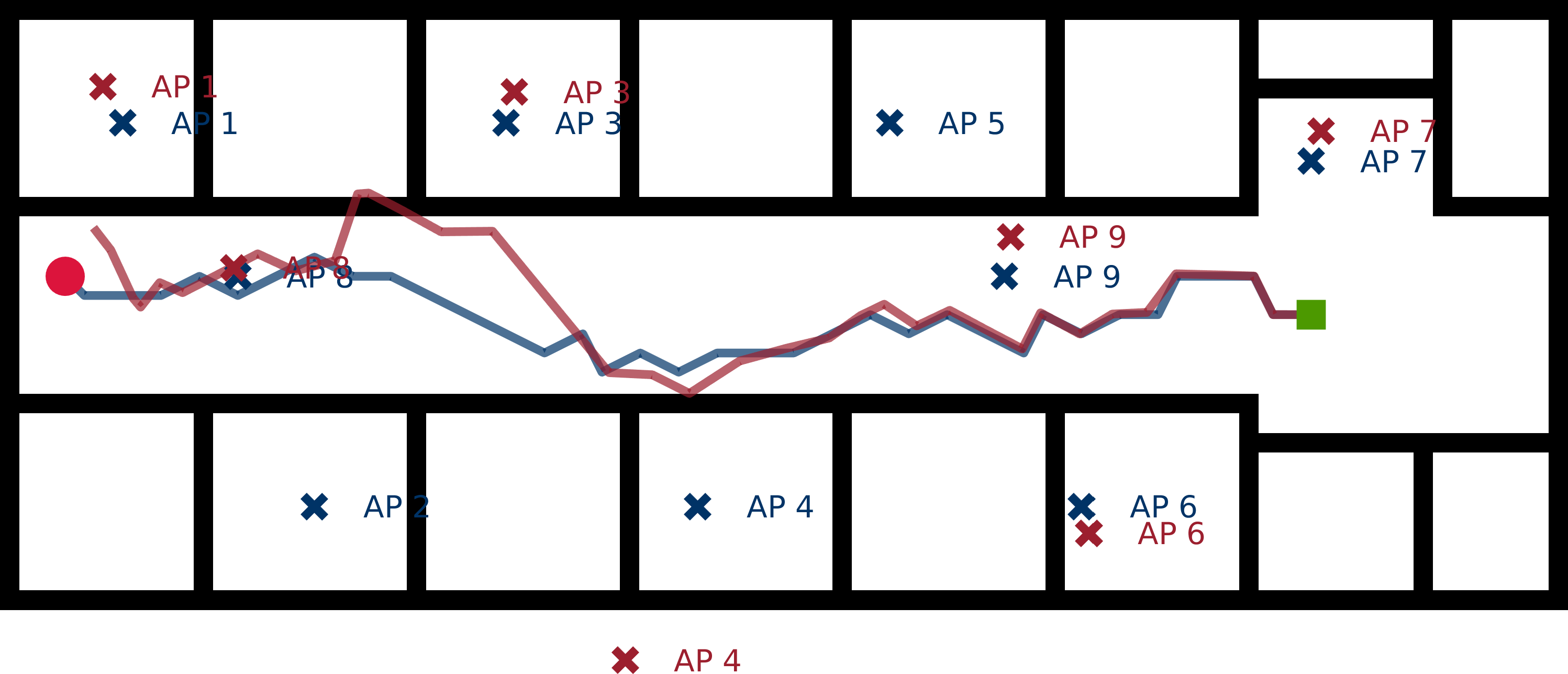}
  \caption{2D Localization in the Office environment using FastSLAM and Soft Classification \textit{(RMSE=0.5258 m)}. No Classification \textit{(RMSE=2.5737 m)} and Hard Classification \textit{(RMSE=0.5377 m)} results are less accurate. All access points have been detected in all 3 cases. The colors show the ground truth as blue and the estimates as red.}
  \label{fig:2dlocfs}
  \squeezeup
\end{figure}

We developed a simulator for testing the performance of the deep network. The random walk algorithm used in the training phase is used to generate ground truth data between a start and goal point. We start simulating the motion of the robot along this path. We then calculate the RSSI value at each of the waypoints in the path, as stated in the previous section. We also calculate the Euclidean distance of the estimated position of the robot from each of the access points (as we know the map within reasonable accuracy). We input these 2 values to the classifier, which then gives us the label (or probabilities) of the signal. We then use this data to update the measurement model accordingly. An example of the localization obtained with the FastSLAM algorithm and soft classification can be seen in Fig. \ref{fig:3dlocfs}. We also attach results for 2D localization in a different environment using FastSLAM and the 3 cases of classifier usage in Fig. \ref{fig:2dlocfs}. We refer the reader to our Github repository for similar results in both 2D and 3D for more complex environments \textit{(such as the Office and W2PTIN)}.

\begin{figure}
  \centering
  \includegraphics[width=0.9\columnwidth,trim={1.4cm 0 1.4cm 0},clip]{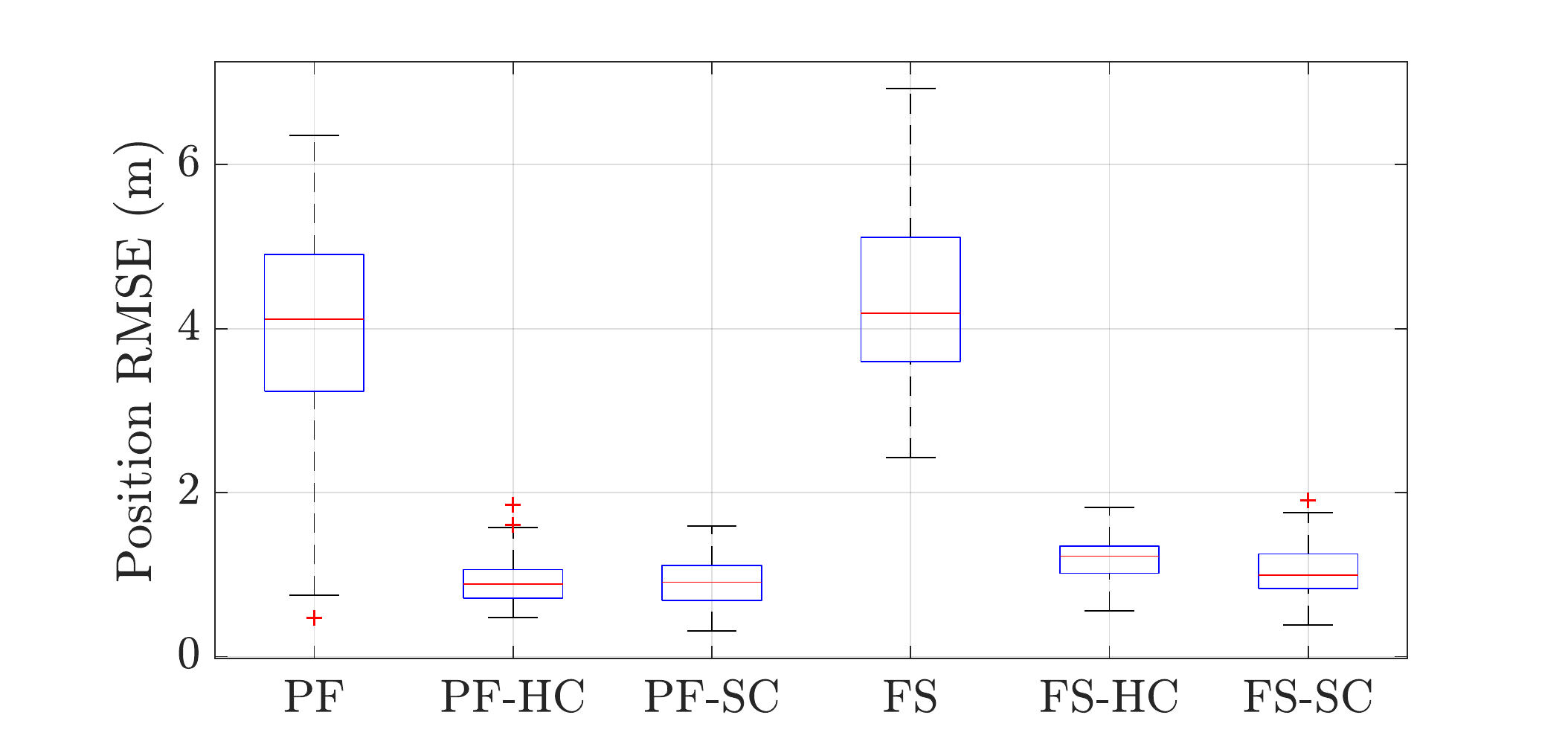}
  \caption{RMSE for 2D localization over 50 trials in the office environment.}
  \label{fig:rmse}
  \squeezeup
\end{figure}

The results obtained in simulation for both the 2D and 3D case for the office environment are given in Table \ref{tab:results}. A set of 50 experiments each was run for each of the scenarios presented in the given table and the mean root mean square error (RMSE) was calculated. As can be seen in Fig. \ref{fig:rmse}, the RMSE shows substantial improvement for localization using both FastSLAM and Particle Filter. 
\par
The parameters used for generating these results are given in Table \ref{tab:params}. The number of particles defines $n_{p}$ used in the particle filter/FastSLAM algorithm. The step size is the length of step taken during the random walk. The motion standard deviation is used to produce white gaussian noise \textit{(zero-centered gaussian distribution with given standard deviation)} \cite{thrun2005probabilistic} in each dimension (x,y,z) which is taken to be a percentage of the step size. The higher the percentage, the noisier the motion model is considered and a more scattered distribution of hypothesis is produced. Similarly, the measurement standard deviation is used to produce white gaussian noise in the range measurements obtained using the RSSI values of access points. Since RSSI based ranging is very inaccurate, we consider the noise to be anywhere between 10m to 100m. In practice, the noise can be higher than 100m, however, in that case, the localization algorithm would totally rely on the motion model for tracking the location rendering the use of a measurement model ineffective. The sensing range defines the maximum range the device is able to sense for WiFi APs. We take the sensing range of the device to be a safe estimate of 15m. Lastly, the AP location noise determines the noise in the location of every AP in each dimension \textit{(only valid for FastSLAM)}.

\begin{figure}
  \centering
  \subfloat[Particle filter]{\includegraphics[width=0.95\columnwidth]{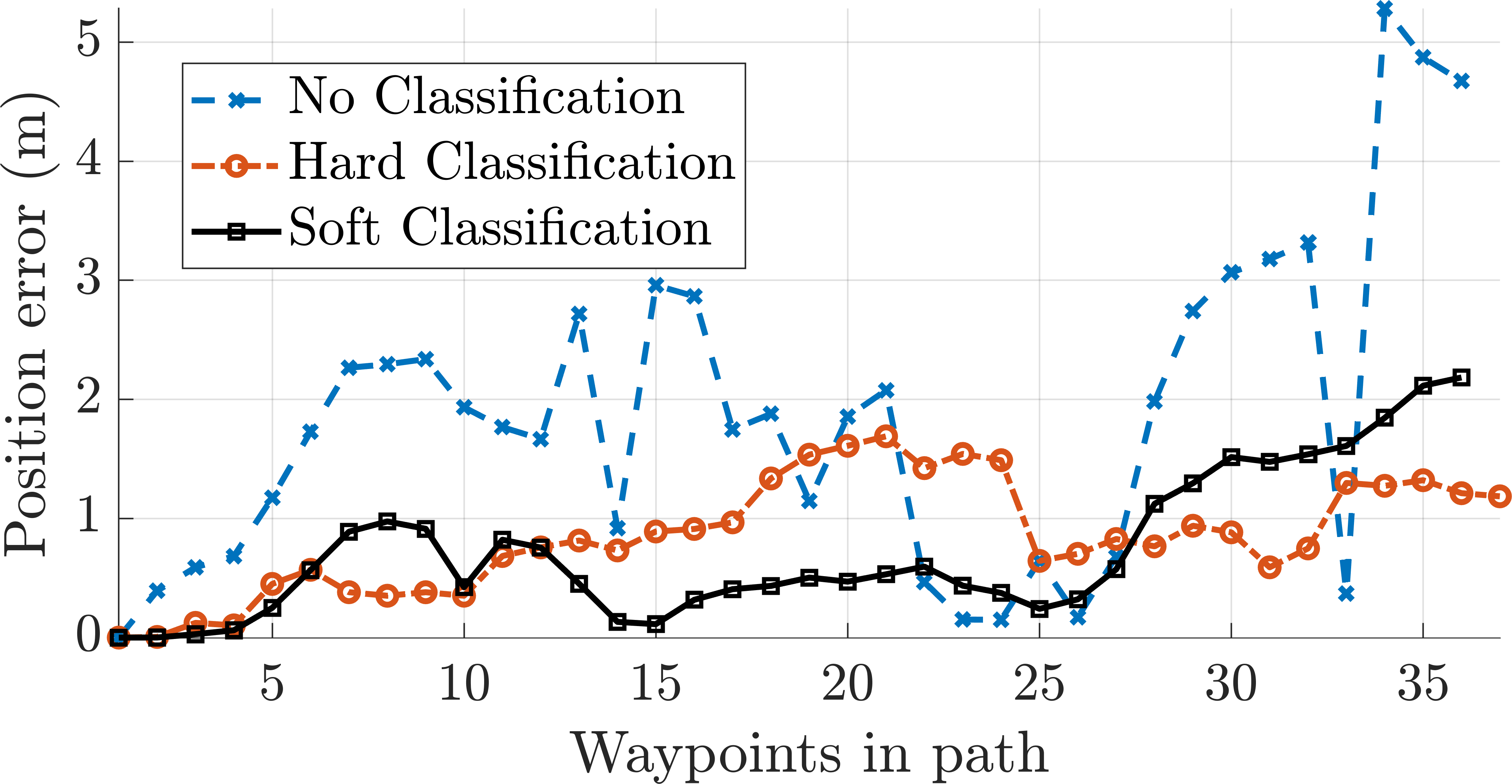}}\\
  \subfloat[FastSLAM]{\includegraphics[width=0.95\columnwidth]{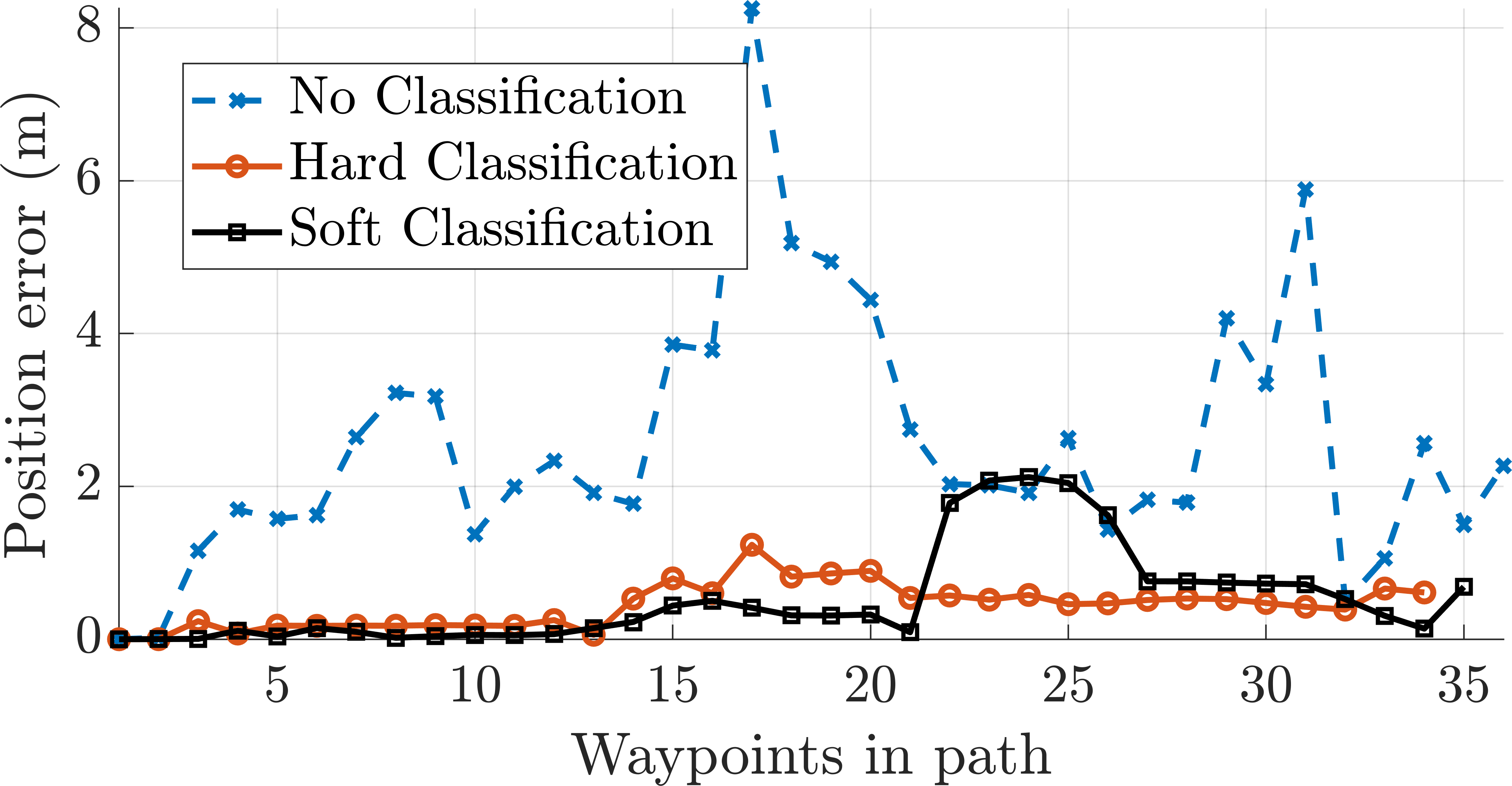}}
  \caption{Error of localized path from ground truth at each waypoint in the Office Environment for a single run using (a) Particle Filter and (b) FastSLAM (corresponds to Fig. \ref{fig:2dlocfs}).}
  \label{fig:errs}
\end{figure}

\begin{table}
    \centering
    \resizebox{\columnwidth}{!}{
    \begin{tabular}{ll}
     \toprule
     \textbf{Parameter Name} &  \textbf{Value}\\
     \midrule
     Number of particles & 3000\\
     Step size & 2m\\
     Motion standard deviation & [rand()*0.8m, rand()*0.8m, rand()*0.8m]\\
     Measurement standard deviation & random value between $[10m-100m]$\\
     Sensing range & 15m\\
     AP location noise & [rand()*10m, rand()*10m, rand()*10m]\\
     \bottomrule
    \end{tabular}
    }
    \caption{Parameters used for Simulation Experiments. Please refer Section \textit{V-B} for details on each parameter. Ignore the last dimension in case of 2D Localization.}
    \label{tab:params}
\end{table}

\subsection{Hardware Experiments}
We used the Fetch Mobile Manipulator for our experiments in one of the buildings of our campus, whose floor plan we had access to. The environment and the robot is as given in Fig. \ref{fig:fetch}. We manually moved the robot around using a remote controller and constructed a map using an open-source implementation of gmapping~\cite{thrun2005probabilistic}. On obtaining the map, we used an open-source implementation of Adaptive Monte Carlo Localization \cite{amcl} for obtaining the ground truth of the robot. We broke down the entire continuous path into waypoints with each segment at least 2.5m apart. At each waypoint, we recorded the WiFi signatures (includes signal strength and MAC address) of all the access points present and the visual odometry calculated using ORB-SLAM ~\cite{mur2015orb}.
\par

Once we gathered all the WiFi, odometry and waypoint information, we ran the FastSLAM algorithm for all three cases. Particle Filter wasn't run as we had a certain measure of ambiguity in the location of all the access points.
Localization is only performed in two dimensions because of the sensing capabilities of the robot.
\par

\begin{figure*}[t!]
  \centering
  \includegraphics[width=1.88\columnwidth]{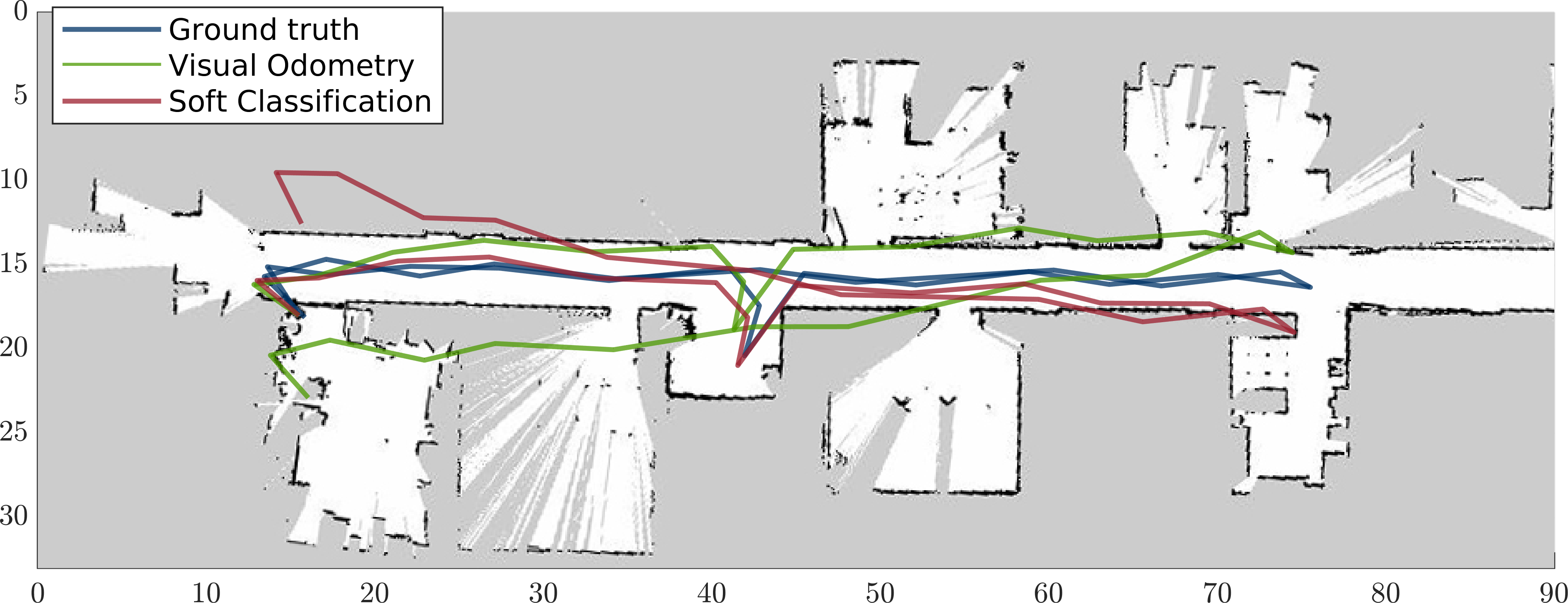}
  \caption{2D Localization in the the University of Michigan Naval Architecture and Marine Engineering (NAME) building (35m x 170m) using FastSLAM and Soft Classification in a single run. The occupancy grid map is only used for visualization.}
  \label{fig:errexpFSSC}
\end{figure*}

We assumed the motion standard deviation of the robot to be $1 m$ in both $x$ and $y$ axes based on experiments in our domain using ORB-SLAM~\cite{mur2015orb}. All other parameters were same as in the case of simulation. Step size and sensing range are not applicable in this case. The access point locations as well weren't accurate as the floor plan wasn't to scale. Hence we assume a certain error ($\leq 0.8 \m$) in their locations as well. The results obtained using FastSLAM for one of our trials is given in Table \ref{tab:expres}. The result for 2D localization using soft classification in FastSLAM can be seen in Fig. \ref{fig:errexpFSSC}.

\begin{table}[t]
\centering
\resizebox{\columnwidth}{!}{
\begin{tabular}{cccc}
    \toprule 
    \multicolumn{4}{c}{\textbf{Indoor localization using the Fetch Mobile Manipulator and FastSLAM (FS)}} \\
    \midrule 
    Alg. &  Classifier (Y/N) & Hard/Soft Class. (H/S) & Loc. RMSE (in m)\\
    \midrule
    FS & N & N/A & 4.3291 \\
    FS & Y & H & 3.6855 \\ 
    \textbf{FS} & \textbf{Y} & S & \textbf{1.6271} \\
    \bottomrule
\end{tabular} }
\caption{Results for Hardware Experiments in the NAME building, Ann Arbor for a single trial. We only use FastSLAM (FS) with the measurement models as described in the DeepLocNet Framework. For each experiment we calculate the RMSE (root mean square error) between each of the waypoints of the ground truth and the localized path.}
\label{tab:expres}
\end{table}

\section{Conclusion and Future Work}
We presented a deep learning-based approach for classifying LOS/NLOS signal propagation, aiding in better localization estimates solely based on signal strength measurements. In particular, the proposed method can provide effective localization using the available infrastructure without the use of any custom hardware. We further incorporated the proposed deep learning-based classifier into the measurement model of particle filtering and FastSLAM. Our experiments show that the resulting system can track a person moving through a large and highly-structured indoor environment with accuracy within $2 \m$. We also present results in indoor environments using signal strength from multiple access points. The presented results in this work show that the DeepLocNet framework performs promisingly better than available wireless-based positioning systems which have an accuracy of $1-10\m$~\cite{liu2007}.

Future work includes modifying the network to incorporate information from the PHY layer as well, further improving the accuracy of localization. However, for this to happen, mobile devices would have to be fitted with the required hardware.

\addtolength{\textheight}{-12cm}   


\section*{Acknowledgment}
{\small 
This work was partially supported by the Toyota Research Institute (TRI), partly under award number N021515, however, this article solely reflects the opinions and conclusions of its authors and not TRI or any other Toyota entity. We also thank Dr. Corina Barbalata and Deep Robot Optical Perception Lab for providing us with the help and support for all hardware experiments.}


\bibliographystyle{IEEEtran}
\bibliography{strings-abrv,ieee-abrv,refs}

\begin{thebibliography}{10}
\providecommand{\url}[1]{#1}
\csname url@rmstyle\endcsname
\providecommand{\newblock}{\relax}
\providecommand{\bibinfo}[2]{#2}
\providecommand\BIBentrySTDinterwordspacing{\spaceskip=0pt\relax}
\providecommand\BIBentryALTinterwordstretchfactor{4}
\providecommand\BIBentryALTinterwordspacing{\spaceskip=\fontdimen2\font plus
\BIBentryALTinterwordstretchfactor\fontdimen3\font minus
  \fontdimen4\font\relax}
\providecommand\BIBforeignlanguage[2]{{%
\expandafter\ifx\csname l@#1\endcsname\relax
\typeout{** WARNING: IEEEtran.bst: No hyphenation pattern has been}%
\typeout{** loaded for the language `#1'. Using the pattern for}%
\typeout{** the default language instead.}%
\else
\language=\csname l@#1\endcsname
\fi
#2}}

\bibitem{amiot2013pylayers}
N.~Amiot, M.~Laaraiedh, and B.~Uguen, ``Pylayers: An open source dynamic
  simulator for indoor propagation and localization,'' in \emph{Communications
  Workshops (ICC), 2013 IEEE International Conference on}.\hskip 1em plus 0.5em
  minus 0.4em\relax IEEE, 2013, pp. 84--88.

\bibitem{pap0}
L.~D. Nardis, ``{3D} indoor positioning and navigation: Theory and
  implementation,'' 2017, {\tt morgner@uni-muenster.de}.

\bibitem{pap1}
\BIBentryALTinterwordspacing
Tektronix. (2014) {Wi-Fi}: Overview of the 802.11 physical layer and
  transmitter measurements. [Online]. Available:
  \url{https://www.cnrood.com/en/media/solutions/Wi-Fi_Overview_of_the_802.11_Physical_Layer.pdf}
\BIBentrySTDinterwordspacing

\bibitem{thrun2005probabilistic}
S.~Thrun, W.~Burgard, and D.~Fox, \emph{Probabilistic robotics}.\hskip 1em plus
  0.5em minus 0.4em\relax MIT press, 2005.

\bibitem{pap9}
P.~Bahl, V.~N. Padmanabhan, \emph{et~al.}, ``Radar: An in-building rf-based
  user location and tracking system,'' in \emph{INFOCOM}, vol.~2, no.
  2000.\hskip 1em plus 0.5em minus 0.4em\relax IEEE, 2000, pp. 775--784.

\bibitem{pap8}
M.~Youssef and A.~Agrawala, ``The horus wlan location determination system,''
  in \emph{Proceedings of the 3rd international conference on Mobile systems,
  applications, and services}.\hskip 1em plus 0.5em minus 0.4em\relax ACM,
  2005, pp. 205--218.

\bibitem{pap19}
X.~He, S.~Badiei, D.~Aloi, and J.~Li, ``Wifi ilocate: Wifi based indoor
  localization for smartphone,'' in \emph{Wireless Telecommunications
  Symposium}.\hskip 1em plus 0.5em minus 0.4em\relax IEEE, 2014, pp. 1--7.

\bibitem{pap16}
J.~L. Carrera, Z.~Li, Z.~Zhao, T.~Braun, and A.~Neto, ``A real-time indoor
  tracking system in smartphones,'' in \emph{Proceedings of the 19th ACM
  International Conference on Modeling, Analysis and Simulation of Wireless and
  Mobile Systems}.\hskip 1em plus 0.5em minus 0.4em\relax ACM, 2016, pp.
  292--301.

\bibitem{jadidi2017gaussian}
M.~Ghaffari~Jadidi, M.~Patel, and J.~Valls~Miro, ``Gaussian processes online
  observation classification for rssi-based low-cost indoor positioning
  systems,'' in \emph{Proc. {IEEE} Int. Conf. Robot. and Automation}.\hskip 1em
  plus 0.5em minus 0.4em\relax IEEE, 2017, pp. 6269--6275.

\bibitem{pap7}
S.~Sen, R.~R. Choudhury, B.~Radunovic, and T.~Minka, ``Precise indoor
  localization using phy layer information,'' in \emph{Proceedings of the 10th
  ACM Workshop on hot topics in networks}.\hskip 1em plus 0.5em minus
  0.4em\relax ACM, 2011, p.~18.

\bibitem{pap5}
Z.~Zhou, Z.~Yang, C.~Wu, L.~Shangguan, H.~Cai, Y.~Liu, and L.~M. Ni,
  ``Wifi-based indoor line-of-sight identification,'' \emph{IEEE Transactions
  on Wireless Communications}, vol.~14, no.~11, pp. 6125--6136, 2015.

\bibitem{pap14}
Z.~Li, Z.~Tian, M.~Zhou, and Y.~Jin, ``Wi-vision: An accurate and robust
  los/nlos identification system using hopkins statistic,'' in \emph{Global
  Communications Conference}.\hskip 1em plus 0.5em minus 0.4em\relax IEEE,
  2017, pp. 1--6.

\bibitem{pap13}
F.~Xiao, Z.~Guo, H.~Zhu, X.~Xie, and R.~Wang, ``Ampn: Real-time los/nlos
  identification with wifi,'' in \emph{IEEE International Conference on
  Communications (ICC)}.\hskip 1em plus 0.5em minus 0.4em\relax IEEE, 2017, pp.
  1--7.

\bibitem{pap15}
X.~Wang, L.~Gao, and S.~Mao, ``Biloc: Bi-modal deep learning for indoor
  localization with commodity 5ghz wifi,'' \emph{IEEE Access}, vol.~5, pp.
  4209--4220, 2017.

\bibitem{pap12}
X.~Wang, X.~Wang, and S.~Mao, ``Deep convolutional neural networks for indoor
  localization with csi images,'' \emph{IEEE Transactions on Network Science
  and Engineering}, 2018.

\bibitem{agarap2018deep}
A.~F. Agarap, ``Deep learning using rectified linear units (relu),''
  \emph{arXiv preprint arXiv:1803.08375}, 2018.

\bibitem{GhaffariJadidi2018}
M.~Ghaffari~Jadidi, M.~Patel, J.~V. Miro, G.~Dissanayake, J.~Biehl, and
  A.~Girgensohn, ``A radio-inertial localization and tracking system with ble
  beacons prior maps,'' in \emph{Proceedings of the IEEE International
  Conference on Indoor Positioning and Indoor Navigation}, 2018, pp. 1--8.

\bibitem{amcl}
D.~Fox, W.~Burgard, F.~Dellaert, and S.~Thrun, ``Monte carlo localization:
  Efficient position estimation for mobile robots,'' \emph{AAAI/IAAI}, vol.
  1999, no. 343-349, pp. 2--2, 1999.

\bibitem{mur2015orb}
R.~Mur-Artal, J.~M.~M. Montiel, and J.~D. Tardos, ``Orb-slam: a versatile and
  accurate monocular slam system,'' \emph{IEEE transactions on robotics},
  vol.~31, no.~5, pp. 1147--1163, 2015.

\bibitem{liu2007}
H.~Liu, H.~Darabi, P.~Banerjee, and J.~Liu, ``Survey of wireless indoor
  positioning techniques and systems,'' \emph{IEEE Trans. on Systems, Man, and
  Cybernetics, Part C}, vol.~37, no.~6, pp. 1067 -- 1080, 2007.

\end{thebibliography}
\end{document}